\documentclass[11pt,a4paper]{article}
\usepackage[hyperref]{acl2021}
\usepackage{nert}
\usepackage{times}
\usepackage{latexsym}

\usepackage{url}
\usepackage[T1]{fontenc}
\usepackage[toc,page]{appendix}
\usepackage{adjustbox}
\usepackage{xr}

\newcommand{\labeledsection}[2]{\section{#1}\label{sec:#2}}

\makeatletter
\newcommand*{\addFileDependency}[1]{% argument=file name and extension
  \typeout{(#1)}
  \@addtofilelist{#1}
  \IfFileExists{#1}{}{\typeout{No file #1.}}
}
\makeatother
 
\newcommand*{\myexternaldocument}[1]{%
    \externaldocument{#1}%
    \addFileDependency{#1.tex}%
    \addFileDependency{#1.aux}%
}

\myexternaldocument{streusletagger-short-appendix}

\usepackage{microtype}

\aclfinalcopy % Uncomment this line for the final submission

\title{Lexical Semantic Recognition}

\author{Nelson F. Liu \\
  Stanford University \\
  \eml{nfliu@cs.stanford.edu} \\
  \And
Daniel Hershcovich \\
  University of Copenhagen \\
   \eml{dh@di.ku.dk}
  \AND
Michael Kranzlein \quad Nathan Schneider\\
  Georgetown University\\
  \{\emldisplay{mmk119@georgetown.edu}{mmk119}, \emldisplay{nathan.schneider@georgetown.edu}{nathan.schneider}\}\texttt{@georgetown.edu}
}

\date{}

\begin{document}
\maketitle
\begin{abstract}
In lexical semantics, full-sentence segmentation and segment labeling of various phenomena are generally treated separately, despite their interdependence.
% \nss{specifically full-sentence lexical segmentation? joint BIO tagging+labeling is not a new idea, but e.g. NER, traditional supersense tagging, or VMWE recognition is narrowly focused}
%We therefore unify several previously-disparate styles of lexical semantic annotation, and
We hypothesize that a unified \textit{lexical semantic recognition} task is an effective way to encapsulate previously disparate styles of annotation, including multiword expression identification\slash classification and supersense tagging.
%We evaluate a standard neural CRF model along all annotation axes available in the STREUSLE corpus, namely lexical unit segmentation (multiword expressions), word-level syntactic tags, and supersense classes for noun, verb, and preposition/possessive units.
Using the STREUSLE corpus, we train a neural CRF sequence tagger and evaluate its performance along various axes of annotation.
As the label set generalizes that of previous tasks (PARSEME, DiMSUM), we additionally evaluate how well the model generalizes to those test sets, finding that it approaches or surpasses existing models despite training only on STREUSLE.
Our work also establishes baseline models and evaluation metrics for integrated and accurate modeling of lexical semantics, facilitating future work in this area.
\end{abstract}

\section{Introduction}\label{sec:introduction}

Many NLP tasks traditionally approached as tagging
%such as named entity recognition, supersense tagging, and multiword expression (MWE) identification, 
focus on lexical semantic behavior---they aim to identify and categorize lexical semantic units in running text using a general set of labels. 
Two examples are supersense tagging of nouns and verbs as formulated by \citet{ciaramita-06}, and verbal multiword expression (MWE) identification and classification in the multilingual PARSEME shared tasks \citep{parseme1.0,parseme1.1,parseme1.2}.
By analogy with named entity recognition, we can use the term \textbf{lexical semantic recognition} (LSR) for such chunking-and-labeling tasks that apply to lexical meaning generally, not just entities.
This disambiguation can serve as a foundational layer of analysis for downstream applications in natural language processing, and provides an initial level of organization for compiling lexical resources, such as semantic nets and thesauri.

In this paper, we tackle a more inclusive LSR task of lexical semantic segmentation and disambiguation.
The STREUSLE corpus (see \Cref{sec:frameworks}) %\footnote{STREUSLE consists of English web reviews, but the style of annotation is not specific to English or the reviews genre.} \citep{schneider-15,schneider-18} 
contains comprehensive annotations of MWEs (along with their holistic syntactic status) and noun, verb, and preposition/possessive supersenses.
We train a neural CRF tagger \cite{Lafferty:2001:CRF:645530.655813} using BERT embeddings \cite{devlin-etal-2019-bert} and find that it obtains strong results %in a multifaceted evaluation (serving 
as a first baseline for this task in its full form.

In addition, we ask: Does a tagger trained on STREUSLE generalize to evaluations like %those featured in 
the PARSEME shared task on verbal MWEs \citep{parseme1.1} and the DiMSUM shared task on MWEs and noun/verb supersenses \citep{schneider-etal-2016-semeval}?
% However, does it strictly generalize all of their distinctions,
% or does it lose some of them for uniformity?\nss{TODO}
% \nss{Frame the core question here. Is it about generalizability? *In principle* STREUSLE's annotation scheme subsumes the others, but we are not sure whether *in practice* this is true?}
%
%To investigate this question,
%, and on individual subtasks as evaluated on both STREUSLE and the more narrowly focused established benchmarks (\Cref{sec:tagger}).
Results show our LSR model based on STREUSLE is general enough to capture different types of analysis consistently, and suggest an integrated full-sentence tagging framework is valuable for explicit modeling of lexical semantics in NLP.\footnote{Code, pretrained models, and model and scorer output (all train/dev/test splits) can be found at \url{https://nelsonliu.me/papers/lexical-semantic-recognition}}

\section{LSR Tagging Frameworks}\label{sec:frameworks}

\begin{figure*}[t]\centering
    \includegraphics[width=0.9\textwidth]{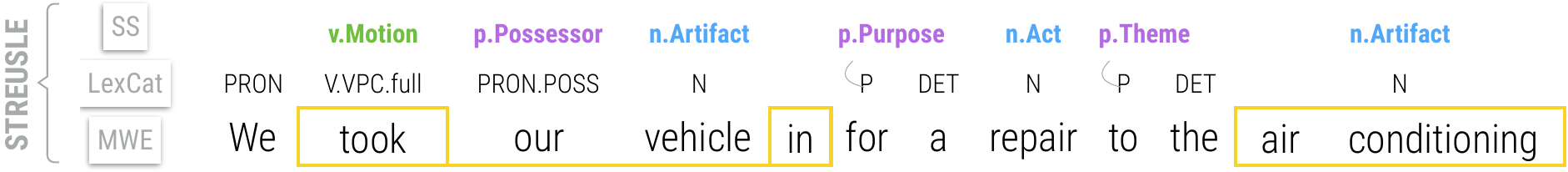}
    \caption{%Example sentence from the Reviews training set
    %(\texttt{reviews-086839-0003}, ``We took our vehicle in for a repair to the air conditioning''),
    %with STREUSLE annotations. 
    Example annotated sentence from the STREUSLE training set.
    The (strong) multiword expressions ``took\dots in'' and ``air conditioning'' each receive a single lexcat and supersense.
    UD syntax is not shown.}
    \label{fig:ex}
\end{figure*}

%In this work, we mainly address the rich lexical semantic analysis in
%STREUSLE (\cref{sec:streusle}),
%but also address DiMSUM and PARSEME (\cref{sec:related_frameworks}).

%\subsection{STREUSLE}\label{sec:streusle}
Our tagger is based on 
STREUSLE \citep[Supersense-Tagged Repository of English with a Unified Semantics for Lexical Expressions;][]{schneider-15,schneider-18},\footnote{\url{https://github.com/nert-nlp/streusle}}
a corpus of web reviews annotated comprehensively for lexical semantic units and supersense labels.
Specifically, there are three~annotation layers: \textbf{multiword expressions}, \textbf{lexical categories}, and \textbf{supersenses}. 
The supersenses apply to noun, verb, and prepositional\slash possessive units.
\Cref{fig:ex} shows an example.

Many of the component annotations have been applied to other languages: verbal multiword expressions \citep{parseme1.0,parseme1.1}, noun and verb supersenses \citep[e.g.,][]{picca-08,qiu-11,schneider-13,martinez_alonso-15,hellwig-17}, and adposition supersenses \citep{hwang-17,zhu-19}.
In this paper we focus on English, where comprehensive annotation is available.

%STREUSLE's annotation layers are detailed in \cref{sec:layers}.
%\Cref{sec:related_frameworks} presents related corpora.

\subsection{STREUSLE Annotation Layers}\label{sec:layers}

STREUSLE comprises the entire 55K-word Reviews section of the English Web Treebank \citep{bies2012english}, for which there are gold Universal Dependencies \cite[UD;][]{nivre-20} graphs, and adopts the same train/dev/test split.

The lexical-level annotations do not make use of the UD parse directly, but there are constraints on compatibility between lexical categories and UPOS tags (see \Cref{sec:constraints}).

\textbf{Multiword expressions} \citep[MWEs;][]{baldwin-10} are expressed as groupings of two or more tokens into idiomatic or collocational units. 
As detailed by \citet{schneider-14-tacl,schneider-14-corpus}, these units may be contiguous or \emph{gappy} (discontinuous).\footnote{The gap in a discontinuous MWE may contain single-word and/or other multiword expressions, provided that those embedded MWEs do not themselves contain gaps.}
Each unit is marked with a binary \emph{strength} value: idiomatic\slash noncompositional expressions are \emph{strong}; collocations that are nevertheless semantically compositional, like ``highly recommended'', are \emph{weak}.

We use the term \textbf{lexical unit} for any expression that is either a \emph{strong} MWE grouping of multiple tokens, or a token that does not belong to a strong MWE.
Every token in the sentence thus belongs to exactly one lexical unit.
The other layers of semantic annotation augment lexical units, 
and weak MWEs are groupings of (entire) lexical units.

\textbf{Lexical categories} (lexcats) describe the syntax of lexical units. 
They are similar to UPOS tags available in the UD annotations of the corpus,
but are necessary in order to (a)~express refinements relevant to the criteria for the application of supersenses, and 
(b)~account for the overall syntactic behavior of strong MWEs, 
which may not be obvious from their internal syntactic structure.\footnote{This is also done in other resources \citep[e.g.,][]{shigeto-13,gerdes-etal-2018-sud}.}
\Cref{sec:lexcat_description} gives the full list of lexcats.
% Lexcats were introduced in version~4.0 of STREUSLE. 
% They were not present in the version of the corpus used by \citet{schneider-15}, and to the best of our knowledge there has been no research on predicting them. 
% However, we expect them to be roughly similar in difficulty to Universal POS tags.

%all kinds of \textbf{multiword expressions} (MWEs) are annotated, giving each sentence a lexical semantic segmentation;
%STREUSLE distinguishes strong MWEs, which are opaque or idiosyncratic in meaning, and weak MWEs, which represent looser collocations.
%syntactic and semantic tags are then applied
%to individual units (single- and multiword). 
%The semantic tags are \textbf{supersenses} for noun, verb, and adpositional\slash possessive units.

\textbf{Supersenses} semantically classify lexical units and provide a measure of disambiguation in context. 
There are 3~sets of supersense labels: 
nominal, verbal, and prepositional\slash possessive.
%\nss{List all the sets in an appendix? see \url{https://github.com/nert-nlp/streusle/blob/master/SUPERSENSES.txt}. I can provide a figure for the SNACS hierarchy of labels}
The lexcat determines which of these sets (if any) should apply.%
%The set of applicable labels---and indeed, whether any supersense should be applied---is determined by the lexcat.
\footnote{\label{fn:scene-function}Some preposition units are labeled with two supersenses drawn from the same label set:
the \textbf{scene role} label represents the semantic role of the prepositional \textit{phrase} marked by the preposition, and \textbf{function} label represents the lexical contribution of the \textit{preposition} in itself \citep{schneider-18}. The scene role and the function are identical by default.}

% The \textbf{lexcat} annotations (syntactic category of lexical unit) is a slight extension to the Universal POS tagset, \footnote{STREUSLE tagset documentation:
% \url{https://github.com/nert-nlp/streusle/blob/master/CONLLULEX.md}}

The MWE, lexcat, and supersense information over lexical units is serialized as per-token tags in a BIO-based encoding (details in \Cref{sec:serialization}).

\subsubsection{Tag Serialization}\label{sec:serialization}

STREUSLE specifies \textbf{token-level tags} to allow modeling lexical semantic recognition as sequence tagging.
The \verb|BbIiOo_~| tagging scheme \cite{schneider-14-tacl} consists of 
8 positional flags indicating MWE status: 
\tg{\flag{O}} applies to single-word expressions, \tg{\flag{B}} to the start of a new MWE, \flag{\tg{I\_}} to the continuation of a strong MWE, and \flag{\tg{I\tat}} to the continuation of a weak MWE (if not continuing a strong MWE within the weak MWE). 
The lowercase counterparts \tg{\flag{o}}, \tg{\flag{b}}, \flag{\tg{i\_}}, \flag{\tg{i\tat}} are the same except they are used within the gap of a discontinuous MWE.
For MWE identification, local constraints on tag bigrams---e.g., that the bigrams $\langle\tg{\flag{B}},\tg{\flag{B}}\rangle$ and $\langle\tg{\flag{B}},\tg{\flag{O}}\rangle$ are invalid, and that the sentence must end with \flag{\tg{I\_}}, \flag{\tg{I\tat}}, or \flag{\tg{O}}---ensure a valid overall segmentation into units \citep{schneider-15}.

The lexcat and (where applicable) supersense information is incorporated in the \emph{first} tag of each lexical unit.\footnote{Though in named entity recognition it is typical to include the class label on every token in the multiword unit, STREUSLE does not do this because it would create a nonlocal constraint across gaps (that the tags at either end have matching lexcat and supersense information). 
A tagger would either need to use a more expensive decoding algorithm or would need to greatly enhance the state space so within-gap tags capture information about the gappy expression.

In STREUSLE there is actually a slight limitation due to the verbal lexcats, which distinguish between single-word and strong multiword expressions (see \Cref{sec:lexcat_description}): if a \tg{\flag{B}-*} or \tg{\flag{I\tat}-*} tag is followed by a gap, there is no local indication of whether the expression will be strong or weak (strength is indicated only after the gap).
If the expression being started is strong, then one of the verbal MWE subtypes (\texttt{V.VID}, etc.)\ should apply; whereas the correct lexcat for a single-word verb is plain \texttt{V}. In practice this is not a problem.}
Thus \tg{\flag{B}-N-n.ARTIFACT} indicates the beginning of an MWE whose lexcat is \tg{N} and supersense is \sst{n.artifact}. 
\flag{\tg{I\_}} and \flag{\tg{i\_}} tags never contain lexcat or supersense information as they continue a lexical unit, whereas \tg{\flag{O}}, \tg{\flag{B}}, \flag{\tg{I\tat}}, \tg{\flag{o}}, \tg{\flag{b}}, and \flag{\tg{i\tat}} always do. 
\Cref{fig:ex_tags} illustrates the full tagging.
All told, STREUSLE has 601 complete tags.

\begin{figure}[th]
    \begin{adjustbox}{frame,minipage=[r][23mm][b]{\columnwidth}}
    We/\tg{\flag{O}-PRON} took/\tg{\flag{B}-V.VPC.full-v.Motion} our/\tg{\flag{o}-PRON.POSS} vehicle/\tg{\flag{o}-N-n.ARTIFACT} in/\tg{\flag{I\_}} for/\tg{\flag{O}-P-p.Purpose} a/\tg{\flag{O}-DET} repair/\tg{\flag{O}-N-n.ACT} to/\tg{\flag{O}-P-p.Theme} the/\tg{\flag{O}-DET} air/\tg{\flag{B}-N-n.ARTIFACT} conditioning/\tg{\flag{I\_}}
    \end{adjustbox}
    \caption{Serialization as token-level tags for the example sentence from \cref{fig:ex}.}
    \label{fig:ex_tags}
\end{figure}

\subsection{Related Frameworks}\label{sec:related_frameworks}\label{sec:parseme}\label{sec:dimsum}

The Universal Semantic Tagset takes a similar approach \citep{bjerva-etal-2016-semantic,abzianidze-17,abdou-etal-2018-learn}, and defines a cross-linguistic inventory of semantic classes for content and function words,
which is designed as a substrate for compositional semantics,
and does not have a trivial mapping to STREUSLE categories.

However, two shared task datasets consist of subsets
of the categories used for STREUSLE annotations, on text from different sources.

\paragraph{PARSEME Verbal MWEs.}
The first such dataset is the English test set for the PARSEME~1.1 Shared Task \citep{parseme1.1},
which covers several genres (including literature and several web genres) and is annotated only for verbal multiword expressions. 
The STREUSLE lexcats for verbal MWEs are identical to those of PARSEME; thus, a tagger that predicts full STREUSLE-style annotations can be evaluated for verbal MWE identification and subtyping by simply discarding the supersenses and the non-verbal MWEs and lexcats from the output.

\paragraph{DiMSUM.}
The second shared task dataset is DiMSUM \cite{schneider-etal-2016-semeval}, 
which was annotated in three genres---TrustPilot web reviews, TED talk transcripts, and tweets---echoing the annotation style of STREUSLE when it contained only MWEs and noun and verb supersenses. DiMSUM does not contain prepositional/possessive supersenses or lexcats. 
It also lacks weak MWEs.
%\nss{we use the test set only; STREUSLE was actually in the training set, along with some other data}

%%%%%%%%%%%%%%%%%%%%%%%%%%%%%%%%%%%%%%%%%%%%%%%%%%%%%%%%%%%%%%%%%%%%%%%%%%%%%%%%%

\section{Modeling}\label{sec:tagger}

\begin{table*}[t]
    \centering\footnotesize\setlength\tabcolsep{3.6pt}
    \begin{tabular}{@{}l|ccc|c|c|ccc||ccc|c@{}}
        %\multirow{3}{*}{All expression sizes}
        \multicolumn{1}{c|}{STREUSLE~4.3} &\multicolumn{3}{c|}{\textbf{Tags}}
        &\textbf{NOUN}&\textbf{VERB}&\multicolumn{3}{c||}{\textbf{SNACS}}&\multicolumn{3}{c|}{\textbf{MWE}} & \textbf{VERB}\\
        \multicolumn{1}{c|}{(test, 5,381 words)} &\textbf{Full}&\textbf{$-$LC}&\textbf{$-$SS}&\textbf{Labeled}&\textbf{Labeled}&\textbf{Labeled}
        &\textbf{Role}&\textbf{Fxn}
        &\multicolumn{3}{c|}{\textbf{LinkAvg}}&\textbf{MWE ID}\\
        &\multicolumn{3}{c|}{\textbf{Accuracy}}&\textbf{F}
        &\textbf{F}&\textbf{F}&\textbf{F}&\textbf{F}
        &\textbf{P}&\textbf{R}&\textbf{F}&\textbf{F}\\
        \hline
        \multicolumn{1}{l|}{\textbf{\# Gold}}&\multicolumn{3}{c|}{5381}&986&697&\multicolumn{3}{c||}{485}&\multicolumn{3}{c|}{433.5}&66\\
        \hline
        BERT {\scriptsize GloVe} (Gold)&82.5 \,\scriptsize 79.3&82.7&89.9&69.0 \,\scriptsize 66.1&77.1 \,\scriptsize 72.1&71.4 \,\scriptsize 61.0&72.4&81.7&80.0&64.9&71.6 \,\scriptsize 59.5&63.9 \,\scriptsize 38.6\\
        BERT {\scriptsize GloVe} (Pred.)&81.0 \,\scriptsize 77.5&81.7&87.9&68.0 \,\scriptsize 65.7&75.1 \,\scriptsize 70.0&71.6 \,\scriptsize 58.0&72.4&82.8&77.6&63.1&69.5 \,\scriptsize 60.3&62.3 \,\scriptsize 43.0\\
        BERT {\scriptsize GloVe} (None)&82.0 \,\scriptsize 77.1	&82.7&89.1&69.6 \,\scriptsize 64.9&76.8 \,\scriptsize 70.3&70.9 \,\scriptsize 58.1	&71.9&81.0&82.0&64.3&72.0 \,\scriptsize 60.3&63.9 \,\scriptsize 42.5\\
        % GloVe (Gold)&79.3&79.5&87.0&66.1&72.1&61.0&61.8&72.5&67.8&53.2&59.5&38.6\\
        % GloVe (Pred.)&77.5&78.1&85.6&65.7&70.0&58.0&59.2&71.3&69.8&53.1&60.3&43.0\\
        % GloVe (None)&77.1&78.0&84.6&64.9&70.3&58.1&59.2&71.5&65.3&56.2&60.3&42.5\\

        \hline
        \citeauthor{schneider-18} &--&--&--&--&--&55.7&58.2&66.7&--&--&--&--
        % \\\\
        % \hline
        % \multirow{2}{*}{MWEs only}
        % &\multicolumn{3}{c|}{\textbf{LinkAvg}}&\multicolumn{3}{H}{\textbf{ID}}
        % &\multicolumn{3}{H|}{\textbf{Labeled}}&\textbf{Labeled}&\textbf{Labeled}&\textbf{Labeled}&
        % \textbf{Role}&\textbf{Fxn}\\
        % &\textbf{P}&\textbf{R}&\textbf{F}&\textbf{P}&\textbf{R}&\textbf{F}&\textbf{P}&\textbf{R}
        % &\textbf{F}&\textbf{F}&\textbf{F}&\textbf{F}&\textbf{F}&\textbf{F}\\
        % \hline
        % \textbf{\# Gold}&\multicolumn{3}{c|}{422}&\multicolumn{6}{H}{284}&155&53&\multicolumn{3}{c}{25}\\
        % \hline
        % BERT (G)&78.5&68.9&73.4&70.9&66.9&68.8&48.9&46.1&47.5&45.8&31.4&62.7&70.6&62.7\\
        % BERT (P)&77.0&68.8&72.7&67.0&65.1&66.1&46.7&45.4&46.1&46.3&30.3&56.6&64.2&60.4\\
        % GloVe (G)&59.5&61.3&60.3&48.3&50.0&49.1&30.6&31.7&31.1&30.7&18.4&44.4&44.4&48.9\\
        % GloVe (P)&66.1&49.1&56.3&52.8&43.3&47.6&33.9&27.8&30.6&29.4&22.7&34.1&43.9&34.1
    \end{tabular}
    \caption{STREUSLE test set results (\%). (Gold): gold POS/lemmas (used in constraints only). (Pred.): predicted POS/lemmas.
    (None): MWE constraints only.
    $-$LC: excluding lexical category. $-$SS: excluding supersense.
    Labeled F: labeled identification F$_1$-score.
    SNACS: preposition supersenses.
    MWE LinkAvg P, R, F: evaluates MWE identification with partial credit.
    Identification of verbal MWEs (exact match) is equivalent to the PARSEME MWE-based metric.
    \citet{schneider-18}: previous best full SNACS tagger, reported on STREUSLE~4.0.
    % \nss{can we get a comparison with Nelson's system (which only does SWE SNACS)?}
    }
    \label{tab:results_streusle_test}
\end{table*}

\begin{table*}[t]
    \centering\footnotesize\setlength\tabcolsep{3.9pt}
    \begin{tabular}{ccc|ccc lr ccc|ccc|cccc}
        \multicolumn{6}{c}{PARSEME 1.1 (\textsc{en}-test, 71,002	words)} & &
        & \multicolumn{10}{c}{DiMSUM 1.0 (test, 16,500 words)} \\
        \cmidrule{1-6}\cmidrule{9-18}
        \multicolumn{3}{c|}{\textbf{MWE-based}}&\multicolumn{3}{c}{\textbf{Token-based}}&
        \multicolumn{2}{c}{ } &\multicolumn{3}{c|}{\textbf{MWEs}}&\multicolumn{3}{c|}{\textbf{Supersenses}}&\multicolumn{4}{c}{\textbf{Combined}}\\
        \textbf{P}&\textbf{R}&\textbf{F}&\textbf{P}&\textbf{R}&\textbf{F}&\multicolumn{2}{c}{ } &\textbf{P}&\textbf{R}&\textbf{F}&\textbf{P}&\textbf{R}&\textbf{F}&\textbf{Acc}&\textbf{P}&\textbf{R}&\textbf{F}\\
        \hline
        \multicolumn{3}{c|}{501}&\multicolumn{3}{c}{1087}&
        \multicolumn{2}{c}{\textbf{\# Gold}}&\multicolumn{3}{c|}{1115}&\multicolumn{3}{c|}{4745}&\multicolumn{4}{c}{5860}\\
        \hline
        36.1&45.5&40.3&40.2&52.0&45.4&
        \multicolumn{2}{c}{BERT  (Gold)}&47.9&52.2&50.0&52.1&56.5&54.2&76.9&51.3&55.7&53.4\\
        34.1&45.9&39.2&37.1&52.2&43.4&
        \multicolumn{2}{c}{BERT  (Pred.)}&48.8&50.7&49.7&49.1&53.9&51.4&75.1&49.1&53.3&51.1\\
        36.2&45.3&40.3&40.4&51.8&45.4&
        \multicolumn{2}{c}{BERT (None)}&53.0&49.2&51.0&50.8&55.1&52.9&76.5&51.2&53.9&52.5\\
        % GloVe (Gold)&27.9&47.4&35.2&39.0&41.6&40.3&67.8&36.0&42.7&39.1\\
        % GloVe (Pred.)&35.8&32.3&34.0&33.4&39.5&36.2&68.3&33.7&38.1&35.8 \\ 
        % GloVe (None)&37.9&37.2&37.5&34.6&40.3&37.2&68.0&35.1&39.7&37.3\\
        \hline
        33.8&32.7&33.3&37.3&31.8&34.4 & Nerima+ & Kirilin+&73.5&48.4&58.4&56.8&59.2&58.0&85.3&59.0&57.2&58.1 \\
        -- & -- & 36.0 & -- & -- & 40.2 & \multicolumn{2}{l}{Taslimipoor+} \\
         -- & -- & 41.9 & -- & -- & -- & \multicolumn{2}{l}{Rohanian+} \\
    \end{tabular}
    \caption{PARSEME and DiMSUM zero-shot test set results (\%) for %The test set covers three domains: reviews, tweets, and TED talks. (G): gold POS/lemmas (used in constraints only). (P): predicted POS/lemmas. 
    BERT models from \cref{tab:results_streusle_test}, compared to prior published results on the tasks.
    GloVe F1 scores (not shown) are 17--20 points below the corresponding BERT scores for PARSEME, and 14--15 for DiMSUM.
    \citet{kirilin-etal-2016-icl}: the best performing system from \citet{schneider-etal-2016-semeval}.
    \Citet{kirilin-etal-2016-icl} and other shared task systems had access to gold POS/lemmas and Twitter training data in addition to all of STREUSLE for training.
    \Citet{nerima-etal-2017-parsing}: a rule-based system which performed best for English in the shared task \citep{parseme1.1}.  \Citet{taslimipoor-19}, \Citet{rohanian-19}: more recent results on the test set (both used ELMo and dependency parses; only some scores were reported).}
    \label{tab:results_dimsum_parseme_combined}
\end{table*}

% \begin{table}[t]
%     \centering\small\setlength\tabcolsep{4pt}
%     \begin{tabular}{@{}l|ccc|ccc@{}}
%         \multicolumn{1}{c|}{PARSEME~1.1} &\multicolumn{3}{c|}{\textbf{MWE-based}}&\multicolumn{3}{c}{\textbf{Token-based}}\\
%         \multicolumn{1}{c|}{VMWEs (\textsc{en}-test)} &\textbf{P}&\textbf{R}&\textbf{F}&\textbf{P}&\textbf{R}&\textbf{F}\\
%         \hline
%         \textbf{\# Gold}&\multicolumn{3}{c|}{501}&\multicolumn{3}{c}{1087}\\
%         \hline
%         BERT (Gold)&36.1&45.5&40.3&40.2&52.0&45.4\\
%         BERT (Pred.)&34.1&45.9&39.2&37.1&52.2&43.4\\
%         BERT (None)&36.2&45.3&40.3&40.4&51.8&45.4\\
%         % GloVe (Gold)&19.9&20.2&20.0&24.8&27.8&26.2\\
%         % GloVe (Pred.)&22.2&22.4&22.3&24.4&25.5&24.9 \\ 
%         % GloVe (None)&21.23&24.2&22.6&25.9&32.4&28.8\\
%         \hline
%         Nerima+ &33.8&32.7&33.3&37.3&31.8&34.4\\
%         Taslimipoor+ & -- & -- & 36.0 & -- & -- & 40.2 \\
%         Rohanian+ & -- & -- & 41.9 & -- & -- & -- \\
%     \end{tabular}
%     \caption{PARSEME test set results for identification (not subtyping) (\%). (Gold): gold POS/lemmas (used in constraints only). (Pred.): predicted POS/lemmas. GloVe F1 scores (not shown) are 17--20 points below the corresponding BERT scores.
%     \Citet{nerima-etal-2017-parsing}: a rule-based system which performed best for English in the shared task \citep{parseme1.1}. \Citet{taslimipoor-19}, \Citet{rohanian-19}: more recent results on the test set (both used ELMo and dependency parses; only some scores were reported).}
%     \label{tab:results_parseme}
% \end{table}

We develop and evaluate a strong neural sequence tagger on the full task of lexical semantic recognition with MWEs and noun/verb/preposition/possessive supersenses to assess the performance of modern techniques on the full joint tagging task. Our tagger feeds pre-trained BERT representations \cite{devlin-etal-2019-bert} through a biLSTM. An affine transformation followed by a linear chain conditional random field produces the final output.
 For further implementation details, see \Cref{sec:model_hyperparameters}.

The predicted tag for each token is the conjunction of its MWE, lexcat, and supersense.\footnote{For prepositions and possessives, the supersense is either a pair of labels, or a single label serving dually as scene role and function (\cref{fn:scene-function}).} There are 572 such tags in the STREUSLE training set, and only 12 unique conjoined tags in the development set are unseen during training ($\approx$5\% of the development set tagging space, corresponding to $\approx$0.2\% of the tokens in the development set). % cut parenthetical if space needed
% These forms of lexical semantics each affect each other---for instance, \nfl{something about how some lexcats can only have some MWEs}? \nfl{Do we have to justify why do we do this, vs. multitask learning? The issue is that MTL with independent tagsets wouldn't pick up all of the constraints between the label spaces, and modeling those interactions is pretty hard. Can punt to future work? Or can just omit.}\oa{There are two issues here: (1) whether we predict a conjoined tag or MTL, (2) whether we enforce constraints between them or let them be learned. For (1), we could say that MTL is future work or not mention it; for (2) we can say that since there are hard constraints on which supersenses are applicable to different lexcats/tags, we enforce them explicitly. }

% A full list of constraints is provided in \cref{sec:model_constraints}.

% \begin{figure}[t]
% \caption{Constraints}\label{fig:model_constraints}
% \end{figure}

\paragraph{Constrained Decoding.}\label{sec:constraints}

%To guarantee valid tagging, we apply several constraints. 
A few hard constraints are imposed in tagging.
To enforce valid \textit{MWE chunks}, we use first-order Viterbi decoding with the appropriate corpus-specific constraints (e.g., for STREUSLE MWEs, the \verb|BbIiOo_~| tagset; see \Cref{sec:serialization}). The MWE constraint is applied during training and evaluation. In addition, a given token's possible lexcats are constrained by the token's \textit{POS tag and lemma}.
For instance, a token with the \verb|AUX| UPOS tag can only take the \verb|AUX| lexcat. However, if the token's UPOS is \verb|AUX| and its lemma is ``\verb|be|'', it can take either the \verb|AUX| or \verb|V| lexcats.

The POS and lemma constraints are only applied during evaluation; to avoid relying on gold POS/lemma annotations at test time we use
%To enforce these constraints, we use 
an off-the-shelf system \cite{qi2018universal}. % to constrain the model to consider only the tags with valid lexcats associated with the predicted UPOS tag and lemma.

\begin{figure*}\small
\resizebox{\textwidth}{!}{
\begin{tabular}{@{}l@{}} % outer tabular
\begin{tabular}{@{}H@{}l@{ }l@{ }l@{ }l@{ }l@{ }l@{ }l@{ }l@{ }l@{ }l@{ }l@{ }l@{ }l@{ }}
Tokens & I & have & a & new & born & daughter & and & she & helped & me & with & a & lot \\
$-$MWE constraints & \textit{O-PRON} & \textit{O-V-v.stative} & \textit{O-DET} & \color{red}\textbf{O-ADJ} & \textit{I\_} & \textit{O-N-n.PERSON} & \textit{O-CCONJ} & \textit{O-PRON} & \textit{O-V-v.social} & \color{red}\textbf{o-PRON} & \textit{O-P-p.Theme} & \textit{B-DET} & \textit{I\_}  \\
$+$MWE = Gold & \textit{O-PRON} & \textit{O-V-v.stative} & \textit{O-DET} & \textit{B-ADJ} & \textit{I\_} & \textit{O-N-n.PERSON} & \textit{O-CCONJ} & \textit{O-PRON} & \textit{O-V-v.social} & \textit{O-PRON} & \textit{O-P-p.Theme} & \textit{B-DET} & \textit{I\_} \\
\end{tabular} \\\midrule
\begin{tabular}{@{}H@{}l@{ }l@{ }l@{ }l@{ }l@{ }l@{ }l@{ }l@{ }l@{ }l@{ }l@{ }l@{ }l@{ }}
Tokens & Go & down & 1 & block & to & Super & 8 & . \\
Tags (+POS and Lemma, no MWE): & \color{red}\textit{B-V.VPC.semi-v.motion} & \color{red}\textit{I\_} & \textit{O-NUM} & \color{red}\textit{O-N-n.COGNITION} & \textit{O-P-p.Goal} & \color{red}\textbf{B-N-n.LOCATION} & \color{red}\textit{O-NUM} & \textit{O-PUNCT} \\
Tags with MWE: & \textit{O-V-v.motion} & \textit{O-P-p.Direction} & \textit{O-NUM} & \color{red}\textit{O-N-n.LOCATION} & \textit{O-P-p.Goal} & \color{red}\textit{B-N-n.LOCATION} & \textit{I\_} & \textit{O-PUNCT} \\
Gold & \textit{O-V-v.motion} & \textit{O-P-p.Direction} & \textit{O-NUM} & \textit{O-N-n.RELATION} & \textit{O-P-p.Goal} & \textit{B-N-n.GROUP} & \textit{I\_} & \textit{O-PUNCT}
\end{tabular} \\\midrule
\begin{tabular}{@{}H@{}l@{ }l@{ }l@{ }l@{ }l@{ }l@{ }l@{ }l@{ }l@{ }l@{ }l@{ }l@{ }l@{ }}
Tokens: & beware & they & will & rip & u & off\\
Tags (+POS and Lemma, no MWE): & \textit{O-V-v.cognition} & \textit{O-PRON} & \textit{O-AUX} & \color{red}\textbf{O-V-v.contact} & \textit{o-PRON} & \textit{I\_} \\
Tags with MWE constraint = Gold: & \textit{O-V-v.cognition} & \textit{O-PRON} & \textit{O-AUX} & \textit{B-V.VPC.full-v.social} & \textit{o-PRON} & \textit{I\_} \\
\end{tabular}
\end{tabular}
}
\caption{Selected examples where the model without MWE constraints (first row under each sentence) produces a structurally invalid tagging.
Incorrect tags are red; the ones that render the tagging structurally invalid are bold.
The last row under each sentence is the gold annotation, and the middle row (if different from gold) is the model prediction with MWE constraints. (The first sentence ends with a period, omitted for brevity.)}
\label{tab:constraint-error-examples}
\end{figure*}

\subsection{Experiments}\label{sec:experiments}

We train the tagger on version 4.3 of the English STREUSLE corpus 
and evaluate on the STREUSLE, English PARSEME, and DiMSUM test sets (\Cref{sec:frameworks}). 
The latter two are (zero-shot) out-of-domain test sets; the tagger is not retrained on the associated shared task training data.

We also compare to a model with static word representations by replacing BERT with the concatenation of 300-dimensional pretrained GloVe embeddings \cite{pennington2014glove} and the output of a character-level convolutional neural network. 
Finally, we also establish an upper bound on performance by providing the model with gold POS tags and lemmas; note that the difference between gold and predicted POS tags and lemmas only applies to the constrained decoding.

\subsection{Results and Discussion}\label{sec:results}

\Cref{tab:results_streusle_test} shows all standard \textbf{STREUSLE} evaluation metrics on the test set.
For preposition supersenses (SNACS), we compare to the results in \citet{schneider-18},
who performed MWE identification and supersense labeling for prepositions only.
Note that \citet{schneider-18} used version 4.0 of the STREUSLE corpus,
which is slightly different from the version we use (some of the SNACS annotations have been revised).
However, our baseline tagger, even with GloVe embeddings,
outperforms \citet{schneider-18} on that subset.
Using BERT embeddings with constraints POS tags and lemmas improves performance substantially; on preposition supersense tagging, it even outperforms using gold POS tags and lemmas.
\citet{liu-etal-2019-linguistic} also found that BERT embeddings improved SNACS labeling on STREUSLE 4.0, although they study a simplified setting (gold preposition identification, and only considering single words).
% \nss{discuss \citep{liu-etal-2019-linguistic} findings w.r.t. SNACS: BERT also helped there, though the setting was easier (gold identification, single-word only). That was STREUSLE 4.0}

\Cref{tab:results_dimsum_parseme_combined} shows standard \textbf{PARSEME} and \textbf{DiMSUM} test set evaluation metrics,
for models trained on the STREUSLE training set, in a zero-shot out-of-domain evaluation setting. On the PARSEME test set, our BERT-based model approaches the state-of-the-art MWE-based F-score and exceeds the best reported \textit{fully-supervised} token-based F-score. %\footnote{It is unclear whether \citet{rohanian-19} used gold syntactic dependencies at test time.}
However, on the DiMSUM test set, the BERT model did not outperform the best shared task system, likely owing to the comparative difficulty of the full lexical semantic recognition task versus the restricted DiMSUM setting.

These results demonstrate that pre-training contextualized embeddings on large corpora
can help models generalize to out-of-domain settings.\footnote{A small fraction of sentences in the PARSEME test set (194$/$3965) are EWT reviews sentences that also appear in STREUSLE's dev set. The rest of the PARSEME test set contains other web and non-web genres \citep{walsh-18}, and thus it is mostly out-of-domain relative to STREUSLE. None of the PARSEME training set overlaps with STREUSLE.}

%We also see that c
% \nss{switching back to STREUSLE?}
Constrained decoding does not substantially impact the performance of our BERT model. In general, constraints with gold POS/lemmas perform the best, while not using POS/lemma constraints is often better than using predicted POS/lemmas. Removing the MWE constraints yields models with slightly higher overall tag accuracy, but results in invalid segmentations for a large proportion of sentences: 14\% of STREUSLE sentences in the fully unconstrained model and 17\% of sentences if only predicted POS and lemmas are used for constraints.

Three sentences out of those 17\% appear in \cref{tab:constraint-error-examples}.
The first shows both an omission of a ``B-'' tag needed to start an MWE (``new'') and a false positive gap without members of an MWE on either side (``me'').
When the full set of constraints is used, the gold tagging is recovered.
In the second sentence, there is a false positive yet structurally valid MWE (``Go down'') as well as an invalid start to an MWE that is never continued (``Super''), perhaps because it is rare for a number to continue an MWE (this happens $<$20 times in the entire corpus).
Finally, in the third sentence, the model constrained only by POS and lemma is inclined toward the literal meaning of ``rip'', whereas the MWE-constrained model recovers the gappy verb-particle construction ``rip off''.
Naturally, in other sentences, the MWE-constrained model sometimes suffers from false positive or false negative MWEs, but always produces a coherent segmentation.

\section{Related Work}\label{sec:related_work}

The computational study of MWEs has a long history
\citep{sag-02,diab-09,baldwin-10,ramisch2015multiword,qu-15,constant-17,bingel-sogaard-2017-identifying,shwartz-dagan-2019-still}, as does supersense tagging \citep{segond-97,ciaramita-06}.
\Citet{vincze-11} developed a sequence tagger for both MWEs and named entities in English.
\Citet{schneider-15,schneider-etal-2016-semeval} featured joint tagging of MWEs and noun and verb supersenses with feature-based sequence models.
\Citet{richardson-17} trained such a model on STREUSLE~3.0 as a noun, verb, and preposition supersense tagger (without modeling MWEs).
%A first-order model proved far superior to a local model 
For preposition supersenses,
% \Citet{bingel-sogaard-2017-identifying} used multitask learning to improve MWE identification and supersense tagging, showing the largest benefits with syntactic chunking as an auxiliary task.
\citet{gonen-16} incorporated multilingual cues;
\citet{schneider-18} experimented with feature-based and neural classifiers; and
\citet{liu-etal-2019-linguistic}, modeling supersense disambiguation of single-word prepositions only, found pretretrained contextual embeddings to be much more effective even with simple linear probing models.

\section{Conclusion}\label{sec:conclusion}

%We propose the task of lexical semantic recognition, which seeks to segment and label units that apply to lexical meaning.
We study the lexical semantic recognition task defined by the STREUSLE corpus, which involves joint MWE identification and coarse-grained (supersense) disambiguation of noun, verb, and preposition expressions; this task subsumes and unifies the previous PARSEME and DiMSUM evaluations.
We develop a strong baseline neural sequence model, and see encouraging results on the task. Furthermore, zero-shot out-of-domain evaluation of our baselines on partial versions of the task yields scores comparable to the fully-supervised in-domain state of the art.

\section*{Acknowledgments}

We are grateful to anonymous reviewers as well as members of the NERT lab for their feedback on this work.
This research was supported in part by NSF award IIS-1812778 and grant 2016375 from the United States--Israel Binational Science Foundation (BSF), Jerusalem, Israel.
NL is supported by an NSF Graduate Research Fellowship under grant number DGE-1656518.

\bibliography{streusletagger-short}
\bibliographystyle{acl_natbib}

\newpage
\clearpage

\onecolumn

\appendix
\begin{appendices}

\labeledsection{Lexical categories in STREUSLE}{lexcat_description}

\begin{table}[h]
    \centering\small
    \begin{tabular}{>{\tt}lcl||>{\tt}ll}
         \multicolumn{1}{l}{\textbf{Lexcat}} & \textbf{SS} & \textbf{Definition} & \multicolumn{1}{l}{\textbf{Lexcat}} & \textbf{Definition} \\
         \hline
         N & n.* & noun, common or proper & NUM & number \\
         \cline{1-3}
         PRON.POSS & p.* & possessive pronoun & PRON & non-possessive pronoun \\
         POSS & p.* & possessive clitic (\textit{'s}) & ADJ & adjective \\
         P & p.* & adposition & ADV & adverb \\
         PP & p.* & (idiomatic) adpositional phrase MWE & DET & determiner \\
         INF.P & p.* & semantically annotatable infinitive marker & INF & nonsemantic infinitive marker \\
         \cline{1-3}
         V & v.* & \textbf{single-word} full verb or copula & AUX & auxiliary, not copula \\
         V.VID & v.* & \textbf{MWE:} verbal idiom & DISC & discourse/pragmatic expression \\
         V.VPC.full & v.* & \textbf{MWE:} full verb-particle construction & CCONJ & coordinating conjunction \\ 
         V.VPC.semi & v.* & \textbf{MWE:} semi verb-particle construction & SCONJ & subordinating conjunction \\ 
         V.LVC.full & v.* & \textbf{MWE:} full light verb construction & INTJ & interjection \\ 
         V.LVC.cause & v.* & \textbf{MWE:} causative light verb construction & SYM & symbol \\
         V.IAV & v.* & \textbf{MWE:} idiomatic adpositional verb & PUNCT & punctuation \\
         & & & X & foreign or nonlinguistic \\
    \end{tabular}
    \caption{Lexcats (lexical categories) that are annotated for strong lexical units, i.e., single-word expressions or strong MWEs. Weak MWEs are treated as compositional and thus do not receive a holistic lextag or supersense.
    \textbf{\textit{Left:}} Lexcats that require supersenses of the class designated in the second column: nominal (n.*), verbal (v.*), or adpositional/possessive (p.*). 
    Verbal MWEs are syntactically subtyped in the lexcat, and the simple \texttt{V} lexcat applies to non-MWEs only.
    \textbf{\textit{Right:}} Lexcats that are incompatible with supersenses. Most of these are defined in line with Universal POS tag definitions, but may also apply to MWEs.
    Definitions come from \url{https://github.com/nert-nlp/streusle/blob/master/CONLLULEX.md}.}
    \label{tab:lexcats}
\end{table}

\labeledsection{Baseline Implementation Details}{model_hyperparameters}

\Cref{table:hyperparameters} lists the hyperparameter values we found
by tuning on the STREUSLE development set, with BERT pre-trained contextualized embeddings \citep[large-cased;][]{devlin-etal-2019-bert},
predicted POS tags and lemmas. BERT parameters are not fine-tuned.

\begin{table}[h]
\centering
\scalebox{.9}{
\begin{tabular}{l|c}
BiLSTM \#layers & 2 \\
BiLSTM total dim. per layer & 512 \\
Learning rate & 0.001 \\
Batch size & 64
\end{tabular}
}
\caption{Hyperparameter values.}
\label{table:hyperparameters}
\end{table}

Our tagger uses the BERT (large, cased) pretrained model to produce input word representations; these input word representations are a learned scalar mixture of the BERT representations, following observations that the topmost layer of BERT is highly attuned to the pretraining task and generalizes poorly \cite{liu-etal-2019-linguistic}. The representation for a token is taken to be BERT output corresponding to its first wordpiece representation. We freeze the BERT representations during training.

The word representations from the frozen BERT contextualizer are then fed into a 2-layer bidirectional LSTM with 256 hidden units in each direction. The LSTM outputs then are projected into the label space with a learned linear function, and a linear chain conditional random field produces the final output.

For training, we minimize the negative log-likelihood of the tag sequence with the Adam optimizer, using a batch size of 64 sequences and a learning rate of 0.001.

We train our model for 75 epochs, and gradient norms are rescaled to a maximum of 5.0. We apply early stopping with a patience of 25 epochs. Our model is implemented in the AllenNLP framework \citep{gardner-etal-2018-allennlp}.

In our ablated models that use GloVe vectors and character-level CNNs instead of BERT, we use 200 output filters with a window size of 5 in the CNN. The input to the CNN are 64-dimensional character embeddings.

\labeledsection{Per-Lexcat STREUSLE Results}{per_lexcat_streusle_results}

\Cref{tab:per_lexcat_streusle_results} shows STREUSLE test set results for the BERT tagger with only MWE constraints (no POS/lemma constraints), broken down by lexical category. The numbers reported here differ from the evaluation in \cref{tab:results_streusle_test}---these metrics are calculated by extracting the predicted and gold spans, and then computing an exact-match F1 measure between the predicted and gold sets.

Frequency counts are for STREUSLE-test; OOV token rates are relative to STREUSLE-train.
Examples are lemmatized lexical units (``lexlemmas'').
Lexlemmas are used to calculate OOV rates.

\begin{table}[h]
    \centering\small\setlength\tabcolsep{3.8pt}
    \begin{tabular}{@{}>{\tt}llcccc||>{\tt}llcccc@{}}
         \multicolumn{1}{l}{\textbf{Lexcat}} & \textbf{Example} & \textbf{\# Gold \scriptsize \% OOV} & \textbf{P} & \textbf{R} & \textbf{F} & \multicolumn{1}{l}{\textbf{Lexcat}} & \textbf{Example} & \textbf{\# Gold \scriptsize \% OOV} &  \textbf{P} & \textbf{R} & \textbf{F}\\
         \midrule
         N & \textit{food} & 946 \,\scriptsize 24.7\% & \hphantom{0}85.5 & \hphantom{0}88.9 & \hphantom{0}87.2 & NUM & \textit{five} & \hphantom{0}41 \,\scriptsize 17.1\% & \hphantom{0}92.9 & 95.1 & 94.0  \\
         \cmidrule{1-6}
         PRON.POSS & \textit{my} & \hphantom{0}94 \,\scriptsize \hphantom{0}0.0\% & \hphantom{0}98.9 & \hphantom{0}93.6 & \hphantom{0}96.2 &  PRON & \textit{it} & 393 \,\scriptsize \hphantom{0}0.0\% & \hphantom{0}95.1 & 98.2 & 96.6  \\
         POSS & \textit{'s} & \hphantom{00}1 \,\scriptsize \hphantom{0}0.0\% & 100.0 & 100.0 & 100.0 &  ADJ & \textit{best} & 532 \,\scriptsize \hphantom{0}8.9\% & \hphantom{0}85.8 & 94.0 & 89.7  \\[3pt]
         P & \textit{with} & 322 \,\scriptsize \hphantom{0}0.3\% & \hphantom{0}88.0 & \hphantom{0}93.2 & \hphantom{0}90.5 &  ADV & \textit{extremely} & 358 \,\scriptsize \hphantom{0}2.1\% & \hphantom{0}91.7 & 92.2 & 91.9  \\
         PP & \textit{by far} & \hphantom{0}18 \,\scriptsize \hphantom{0}0.0\% & \hphantom{0}87.5 & \hphantom{0}77.8 & \hphantom{0}82.4 &  DET & \textit{the} & 376 \,\scriptsize \hphantom{0}0.0\% & \hphantom{0}92.4 & 96.8 & 94.5  \\
         INF.P & \textit{to} see & \hphantom{0}20 \,\scriptsize \hphantom{0}0.0\% & \hphantom{0}87.0 & 100.0 & \hphantom{0}93.0 &  INF & \textit{to} & \hphantom{0}36 \,\scriptsize \hphantom{0}0.0\% & \hphantom{0}91.9 & 94.4 & 93.2  \\
         \cmidrule{1-6}
         V & \textit{go} & 587 \,\scriptsize \hphantom{0}3.5\% & \hphantom{0}90.0 & \hphantom{0}95.4 & \hphantom{0}92.6 &  AUX & \textit{have} & 160 \,\scriptsize \hphantom{0}0.0\% & \hphantom{0}95.7 & 96.2 & 96.0 \\
         V.VID & \textit{take time} & \hphantom{0}24 \,\scriptsize \hphantom{0}0.0\% & \hphantom{0}64.3 & \hphantom{0}37.5 & \hphantom{0}47.4 & DISC & \textit{thanks} & \hphantom{0}21 \,\scriptsize \hphantom{0}4.5\% & \hphantom{0}63.6 & 66.7 & 65.1 \\[3pt]
         V.VPC.full & \textit{turn out} & \hphantom{0}11 \,\scriptsize \hphantom{0}0.0\% & \hphantom{0}58.3 & \hphantom{0}63.6 & \hphantom{0}60.9 &  CCONJ & \textit{and} & 204 \,\scriptsize \hphantom{0}0.0\% & \hphantom{0}95.3 & 99.5 & 97.4  \\ 
         V.VPC.semi & \textit{add on} & \hphantom{00}5 \,\scriptsize \hphantom{0}0.0\% & \hphantom{0}50.0 & \hphantom{0}60.0 & \hphantom{0}54.5 &  SCONJ & \textit{lest} & \hphantom{0}21 \,\scriptsize \hphantom{0}4.8\% & \hphantom{0}90.5 & 90.5 & 90.5 \\[3pt] 
         V.LVC.full & \textit{have fun} & \hphantom{00}8 \,\scriptsize \hphantom{0}0.0\% & \hphantom{0}60.0 & \hphantom{0}37.5 & \hphantom{0}46.2 &  INTJ & \textit{hey} & \hphantom{0}17 \,\scriptsize 35.3\% & \hphantom{0}78.6 & 64.7 & 71.0  \\ 
         V.LVC.cause & \textit{give chance} & \hphantom{00}1 \,\scriptsize \hphantom{0}0.0\% & \hphantom{00}0.0 & \hphantom{00}0.0 & \hphantom{00}0.0 &  SYM & \textit{:)} & \hphantom{0}12 \,\scriptsize \hphantom{0}0.0\% & 100.0 & 75.0 & 85.7 \\[3pt]
         V.IAV & \textit{treat to} & \hphantom{0}17 \,\scriptsize \hphantom{0}5.9\% & \hphantom{0}81.8 & \hphantom{0}52.9 & \hphantom{0}64.3 &  PUNCT & \textit{.} & 597 \,\scriptsize \hphantom{0}0.3\% & \hphantom{0}99.0 & 99.7 & 99.3 \\
         & & & & & & X & \textit{etc} & \hphantom{00}2 \,\scriptsize 50.0\% & \hphantom{00}0.0 & \hphantom{0}0.0 & \hphantom{0}0.0 \\
    \end{tabular}
    \caption{STREUSLE test set results for the BERT-based tagger with only MWE constraints (no POS/lemma constraints), broken down by lexcat. The numbers reported here differ from the evaluation in \cref{tab:results_streusle_test}---these metrics are calculated by extracting the predicted and gold spans, and then computing an exact-match F1 measure between the predicted and gold sets.}
    \label{tab:per_lexcat_streusle_results}
\end{table}

\labeledsection{Per-VMWE Category PARSEME Results}{per_vmwe_category_parseme_results}

\Cref{tab:per_vmwe_category_parseme_results} shows PARSEME (English) test set results for the BERT tagger with only MWE constraints (no POS/lemma constraints), broken down by VMWE category.

Frequency counts are for PARSEME-\textsc{en}-test and reflect the number of gold MWEs; OOV token rates are relative to STREUSLE-train.
Examples are lemmatized lexical units (``lexlemmas'').
Lexlemmas are used to calculate OOV rates.

\begin{table}[h]
    \centering\small\setlength\tabcolsep{4pt}
    \begin{tabular}{>{\tt}llcHccc|ccc}
        \multicolumn{1}{c}{PARSEME~1.1} & \multicolumn{3}{c}{} &\multicolumn{3}{c|}{\textbf{MWE-based}}&\multicolumn{3}{c}{\textbf{Token-based}}\\
        \multicolumn{1}{c}{VMWEs (\textsc{en}-test)} & \textbf{Example} & \textbf{\# Gold \scriptsize \% OOV} & \textbf{\% Gappy} &\textbf{P}&\textbf{R}&\textbf{F}&\textbf{P}&\textbf{R}&\textbf{F}\\
        \midrule
        V.VID & \textit{tide turn} & \hphantom{0}79 \,\scriptsize \hphantom{0}80.6\% & ??.?\% &\hphantom{0}8.8&17.7&11.8&11.8&20.9&15.1\\[3pt]
        
        V.VPC.full & \textit{bring in} & 146 \,\scriptsize \hphantom{0}44.3\% & ??.?\% &41.8&59.6&49.2&43.3&63.7&51.5\\
        V.VPC.semi & \textit{speak up} & \hphantom{0}26 \,\scriptsize \hphantom{0}61.5\% & ??.?\% &12.7&30.8&18.0&12.5&30.8&17.8\\[3pt]
        
        V.LVC.full & \textit{make promise} & 166 \,\scriptsize \hphantom{0}90.5\% & ??.?\% &30.8&\hphantom{0}4.8&\hphantom{0}8.3&38.2&\hphantom{0}6.1&10.6\\
        V.LVC.cause & \textit{yield result} & \hphantom{0}36 \,\scriptsize 100.0\% & ??.?\% &\hphantom{0}0.0&\hphantom{0}0.0&\hphantom{0}0.0&\hphantom{0}0.0&\hphantom{0}0.0&\hphantom{0}0.0\\[3pt]
        
        V.IAV & \textit{turn into} & \hphantom{0}44 \,\scriptsize \hphantom{0}52.2\% & ??.?\% &30.4&38.6&34.0&29.1&37.8&32.9\\
        V.MVC & \textit{cross examine} & \hphantom{00}4 \,\scriptsize \hphantom{0}80.0\% & ??.?\% &\hphantom{0}0.0&\hphantom{0}0.0&\hphantom{0}0.0&\hphantom{0}0.0&\hphantom{0}0.0&\hphantom{0}0.0\\
    \end{tabular}
    \caption{PARSEME (English) test set results for the BERT tagger with only MWE constraints (no POS/lemma constraints), broken down by VMWE category.}
    \label{tab:per_vmwe_category_parseme_results}
\end{table}

\labeledsection{VMWE Performance in STREUSLE vs.~PARSEME}{vmwe_comparison}

PARSEME appears to be a much more challenging task, even considering just the VMWE identification performance in STREUSLE: in the main text, compare BERT model F-scores of 64\% in STREUSLE versus 40\% in PARSEME (where the state-of-the-art result is 42\%).
Why is this the case? We suspect at least three factors are at play:
\begin{itemize}
    \item Substantial domain shift: PARSEME covers a wide range of genres, including literary genres, which is likely to contribute to lower precision and recall in general.
    \item Base rate of MWEs: STREUSLE contains about 10 times as many MWEs per word as PARSEME, in part due to the comprehensive nature of MWE annotation in STREUSLE. Considering just verbal MWEs per word, STREUSLE-train has 763$/$44,815 = 1.7\% and STREUSLE-test has 66$/$5,381 = 1.2\%, whereas PARSEME-test has 501$/$71,002 = 0.7\%.
    So it is not surprising that the STREUSLE-trained tagger would overpredict VMWEs in PARSEME.
    Note that precision is lower than recall overall and for most VMWE subtypes.
    \item OOV rate: MWE identification of lexical items unseen in training is generally more challenging. We see above that the VMWE vocabularies of STREUSLE and PARSEME are largely disjoint, with OOV rates above 50\% for most subtypes. This would be expected to mainly impact recall, and in fact, the higher the OOV rate, in general the lower the recall.
    In particular, recall is much lower than precision for the LVC.full subtype, with an OOV rate of 90.5\%, suggesting that it is able to correctly identify some known LVCs but unable to generalize to new ones. The 8 instances correctly identified had, in fact, been seen in training.
\end{itemize}
\end{appendices}
\end{document}